%% file: main.tex
\documentclass[11pt,a4paper]{article}

\usepackage[utf8]{inputenc}
\usepackage[T1]{fontenc}
\usepackage{amsmath,amssymb}
\usepackage{graphicx}
\usepackage{url}
\usepackage{hyperref}
\usepackage{booktabs}
\usepackage{array}
\usepackage{xcolor}
\usepackage{geometry}
\usepackage{fancyvrb}
\usepackage{listings}
\usepackage{authblk}

\hypersetup{
    colorlinks=true,
    linkcolor=blue,
    citecolor=blue,
    urlcolor=blue,
    breaklinks=true
}

\usepackage{xurl}

\input{metadata.tex}

\begin{document}

\maketitle

\begin{abstract}
	Emergency speech recognition systems exhibit systematic performance degradation on non-standard English varieties, creating a critical gap in services for Caribbean populations. We present TRIDENT (\textbf{T}ranscription and \textbf{R}outing \textbf{I}ntelligence for \textbf{D}ispatcher-\textbf{E}mpowered \textbf{N}ational \textbf{T}riage), a three-layer dispatcher-support architecture designed to structure emergency call inputs for human application of established triage protocols (the ESI for routine operations and START for mass casualty events), even when automatic speech recognition fails.

	The system combines Caribbean-accent-tuned ASR, local entity extraction via large language models, and bio-acoustic distress detection to provide dispatchers with three complementary signals: transcription confidence, structured clinical entities, and vocal stress indicators. \textbf{Our key insight is that low ASR confidence, rather than representing system failure, serves as a valuable queue prioritization signal---particularly when combined with elevated vocal distress markers indicating a caller in crisis whose speech may have shifted toward basilectal registers.} A complementary insight drives the entity extraction layer: trained responders and composed bystanders may report life-threatening emergencies without elevated vocal stress, requiring semantic analysis to capture clinical indicators that paralinguistic features miss.

	We describe the architectural design, theoretical grounding in psycholinguistic research on stress-induced code-switching, and deployment considerations for offline operation during disaster scenarios. This work establishes a framework for accent-resilient emergency AI that ensures Caribbean voices receive equitable access to established national triage protocols. Empirical validation on Caribbean emergency calls remains future work.
\end{abstract}


\vspace{0.5cm}

\input{sections/01_introduction.tex}
\input{sections/02_related_work.tex}
\input{sections/04_theoretical_foundations.tex}
\input{sections/03_system_architecture.tex}
\input{sections/05_deployment.tex}
\input{sections/06_limitations.tex}
\input{sections/07_conclusion.tex}

\bibliographystyle{plain}
\bibliography{refs/references}

\appendix
\input{appendices/implementation.tex}
\input{appendices/acknowledgments.tex}

\end{document}

%% file: metadata.tex
\title{TRIDENT: A Redundant Architecture for Caribbean-Accented Emergency Speech Triage}

\author{Galbraith, E.}
\author{Sutherland, C.}
\author{Morgan, D.}
\affil{SMG Labs Research Group}

\date{\today}

%% file: sections/01_introduction.tex
\section{Introduction}

When a caller dials emergency services during a crisis, modern automatic speech recognition (ASR) systems exhibit well-documented performance disparities across demographic groups \cite{koenecke2020}. For Caribbean English speakers---a population of over 40 million---these disparities compound with a linguistic phenomenon: under acute stress, speakers tend to shift toward basilectal (more creole-heavy) speech registers, precisely the varieties on which ASR systems perform worst.

Caribbean health ministries have adopted internationally-validated triage protocols: the Emergency Severity Index (ESI) for routine operations and START (Simple Triage and Rapid Treatment) for mass casualty events. These protocols assume dispatchers can accurately capture caller information---an assumption that fails systematically when ASR systems cannot reliably transcribe Caribbean speech.

\subsection{TRIDENT: Dispatcher-Empowered Architecture}

\textbf{This paper presents TRIDENT (\textbf{T}ranscription and \textbf{R}outing \textbf{I}ntelligence for \textbf{D}ispatcher-\textbf{E}mpowered \textbf{N}ational \textbf{T}riage)}, designed to ensure Caribbean-accented callers receive equitable access to established triage protocols. Rather than attempting to eliminate ASR errors---an unrealistic goal---we build a \textbf{dispatcher-support system} that remains functional when transcription fails.

\textbf{Our central contribution is a three-layer framework} providing dispatchers with structured inputs for protocol application:

\begin{enumerate}
    \item \textbf{Transcription confidence:} Flags unreliable transcripts so dispatchers know to listen directly to audio

    \item \textbf{Structured entity extraction:} Extracts clinical indicators (location, mechanism, breathing status, vulnerable populations) even from degraded transcriptions

    \item \textbf{Bio-acoustic distress detection:} Provides physiological stress markers independent of transcript content
\end{enumerate}

\subsection{Key Insights}
\label{sec:key_insights}

Two complementary insights motivate this design:

\begin{enumerate}
    \item \textbf{Content beyond voice:} Trained responders and composed bystanders may report life-threatening emergencies without elevated vocal stress. Semantic extraction captures information that paralinguistic features miss---ensuring ``children trapped in burning building,'' spoken calmly, provides dispatchers with structured data for appropriate triage classification.

    \item \textbf{Uncertainty as prioritization signal:} Low ASR confidence, rather than representing failure, serves as a queue prioritization indicator---particularly when combined with elevated vocal distress marking a caller in crisis whose speech may have shifted toward basilectal registers. This reframes accent-induced transcription errors from bugs into features correlating with genuine distress.
\end{enumerate}

TRIDENT addresses critical gaps in existing emergency AI---cloud dependency with accent-agnostic ASR, text-only analysis ignoring paralinguistic signals, dialect blindness to stress-induced register shifting, and infrastructure fragility during disasters---while \textbf{respecting the clinical authority of established protocols}. The system structures inputs and prioritizes queues, but triage decisions remain with trained professionals applying Ministry of Health-mandated frameworks.

%% file: sections/02_related_work.tex
\section{Related Work}

TRIDENT's dispatcher-support architecture draws on research across five domains: ASR for Caribbean varieties, AI in emergency dispatch, vocal stress detection, dialect reversion under cognitive load, and edge computing for disaster resilience.

\input{subsections/related_work/accent_gap.tex}
\input{subsections/related_work/emergency_ai.tex}
\input{subsections/related_work/vocal_stress.tex}
\input{subsections/related_work/dialect_reversion.tex}
\input{subsections/related_work/edge_computing.tex}
\input{subsections/related_work/positioning.tex}

%% file: subsections/related_work/accent_gap.tex
\subsection{The Accent Gap in Automatic Speech Recognition}

Modern ASR systems exhibit systematic performance degradation on non-standard English varieties. Koenecke et al. \cite{koenecke2020} evaluated five commercial ASR systems, finding word error rates averaged 0.35 for Black speakers compared to 0.19 for White speakers, with performance gaps traced to acoustic model limitations rather than language models.

Caribbean English remains especially underserved. Madden et al. \cite{madden2025} developed the first substantial Jamaican Patois corpus (42.58 hours) and derived scaling laws for Whisper performance. Pre-trained Whisper Large achieved 89\% WER on Patois, while fine-tuned Whisper Medium reduced this to 30\% WER. Critically, their scaling law (WER = 158.06 $\times$ M$^{-0.255}$ $\times$ D$^{-0.269}$) demonstrates that dataset increases yield greater gains than model scaling for underrepresented varieties, informing our choice of Whisper Medium with Caribbean-specific fine-tuning.

%% file: subsections/related_work/emergency_ai.tex
\subsection{AI-Assisted Emergency Dispatch and Clinical Protocols}

Emergency services worldwide are exploring AI to improve call handling, but these systems must support established clinical triage protocols rather than replace human judgment.

\textbf{Clinical Triage Protocols.} The Emergency Severity Index (ESI) is a five-level acuity scale (Level 1: immediate lifesaving intervention to Level 5: no resources needed) widely used in the United States and internationally \cite{esi_handbook}. Jamaica's Ministry of Health implemented ESI across all 19 public hospital emergency departments in 2016 \cite{french2020}. For mass casualty events such as hurricanes, the START (Simple Triage and Rapid Treatment) protocol provides rapid four-category sorting: BLACK (deceased/expectant), RED (immediate), YELLOW (delayed), and GREEN (walking wounded). The ESI handbook explicitly notes that ESI should not be used during mass casualty incidents \cite{esi_handbook}.

\textbf{Current AI Systems.} Existing emergency AI systems (e.g., ECA \cite{attiah2025}, Corti \cite{blomberg2019}) achieve promising classification accuracy but rely on cloud-dependent, accent-agnostic ASR and process only transcribed text, ignoring paralinguistic signals. A scoping review of 106 AI studies in prehospital care identified underutilization of multimodal inputs and absence of infrastructure-independent systems as key gaps \cite{chee2023}.

\textbf{Gaps for Caribbean Deployment.} Three limitations motivate TRIDENT's design: (1) no accent adaptation for Caribbean varieties or stress-induced register shifting, (2) no integration of vocal stress detection with text classification, and (3) cloud dependency that fails during disasters when emergency services are most needed. TRIDENT addresses these gaps while maintaining the principle that AI should empower dispatchers to apply ESI/START protocols more effectively, not replace clinical judgment.

%% file: subsections/related_work/vocal_stress.tex
\subsection{Vocal Stress Detection}
\label{sec:vocal_stress}

The bio-acoustic layer builds on research establishing acoustic correlates of psychological stress. A systematic review of 38 studies found fundamental frequency (F0) as the most consistent stress marker, with 15 of 19 studies reporting significant mean F0 increases under stress \cite{schmalz2025}.

Research on emergency communications provides direct validation. Van Puyvelde et al. \cite{vanpuyvelde2018} analyzed real-life emergency recordings including cockpit voice recorders and 911 calls, documenting F0 increases from 123.9 Hz to 200.1 Hz during life-threatening emergencies---a 62\% increase. However, Deschamps-Berger et al. \cite{deschampsberger2021} found that while benchmark IEMOCAP data yielded 63\% emotion recognition accuracy, real emergency calls achieved only 45.6\%---a substantial domain shift. This finding reinforces our design decision to use bio-acoustic analysis as a triage signal routing high-distress calls to human dispatchers, rather than attempting fully automated classification.

%% file: subsections/related_work/dialect_reversion.tex
\subsection{Dialect Reversion Under Cognitive Load}

Psycholinguistic research establishes that for Caribbean speakers navigating the creole continuum---from basilect (most creole features) through mesolect to acrolect (Standard English)---maintaining acrolectal speech requires sustained executive function. The inhibitory control model establishes that non-target languages remain continuously active and must be suppressed through cognitive effort \cite{green1998}. Under high cognitive load, this inhibition fails, causing speakers to revert toward their dominant variety.

Patrick's \cite{patrick1999} sociolinguistic analysis of the Jamaican Creole continuum establishes that stress levels influence speakers' positioning on this spectrum, with most speakers being mesolectal under normal conditions but capable of shifting toward either pole. The implications for emergency services are significant: a professional who speaks Standard English at work may revert toward basilectal Patois when their house is flooding. Standard ASR systems will exhibit precisely the performance degradation documented in the accent gap literature at the moment when accurate recognition is most critical.

%% file: subsections/related_work/edge_computing.tex
\subsection{Edge Computing for Disaster Resilience}

Infrastructure failure during disasters makes the case for offline-capable emergency AI. Hurricane Maria's impact on Puerto Rico saw 95\% of cell towers fail, with the entire island losing power \cite{santosburgoa2020}. Communication infrastructure failure contributed to a disputed death toll ultimately estimated at approximately 3,000, with recovery requiring over 200 days for full power restoration.

Recent model compression advances make edge deployment feasible. Pre-positioned edge computing resources at hospitals, shelters, and emergency coordination centers, loaded with Caribbean-tuned models, could maintain triage capability even during complete grid and network failure.

%% file: subsections/related_work/positioning.tex
\subsection{Summary: Positioning Our Contribution}

TRIDENT addresses four critical gaps in existing emergency dispatch AI for Caribbean deployment:

\begin{itemize}
    \item \textbf{Caribbean-adapted ASR:} Fine-tuned Whisper models (informed by Madden et al.'s scaling laws) provide transcription accuracy for Caribbean speech varieties, enabling viable downstream entity extraction.

    \item \textbf{Multimodal distress detection:} Parallel bio-acoustic analysis provides a signal pathway that functions even when ASR fails, transforming low transcription confidence from a limitation into a queue prioritization feature.

    \item \textbf{Stress-aware design:} Accounts for stress-induced register shifting along the creole continuum---routing calls with elevated vocal distress and low ASR confidence to immediate human attention.

    \item \textbf{Offline operation:} Complete system deployment on edge hardware (Raspberry Pi 5) enables function during infrastructure failures when emergency services are most critical.
\end{itemize}

The result is the first dispatcher-support system designed specifically for Caribbean emergency services---not to make triage decisions, but to ensure Caribbean-accented callers receive equitable access to the ESI and START protocols that their health ministries have adopted. TRIDENT empowers dispatchers with better information and intelligent queue prioritization; clinical judgment remains with trained human professionals.

%% file: sections/04_theoretical_foundations.tex
\section{Theoretical Foundations}

Fine-tuning Whisper on Caribbean speech improves transcription but cannot eliminate the accent gap. Madden et al. \cite{madden2025} achieved 30\% WER on Jamaican Patois---dramatic improvement from 89\% baseline, but still far above the $<$5\% WER typical for standard English. Moreover, fine-tuning on broadcast speech cannot capture emergency acoustics: elevated noise, emotional qualities, and stress-induced basilectal reversion. ASR alone will fail when needed most.

Conversely, bio-acoustic distress detection cannot provide semantic information needed for dispatch. A caller may exhibit extreme vocal stress while saying ``my house is on fire'' or ``I lost my keys''---identical distress signals but dramatically different responses. Furthermore, Deschamps-Berger et al. \cite{deschampsberger2021} found laboratory emotion recognition accuracy (63\%) drops substantially in real emergency calls (45.6\%). Bio-acoustic features provide gradient information about caller state but cannot substitute for semantic content.

\subsection{The Integration Thesis}

Our architecture integrates these complementary information sources based on the following thesis: \textbf{In emergency contexts, the correlation between ASR failure and genuine distress creates an opportunity to use recognition uncertainty as a routing signal rather than an error to be minimized.}

\begin{figure}[!htb]
\centering
\includegraphics[width=0.85\textwidth]{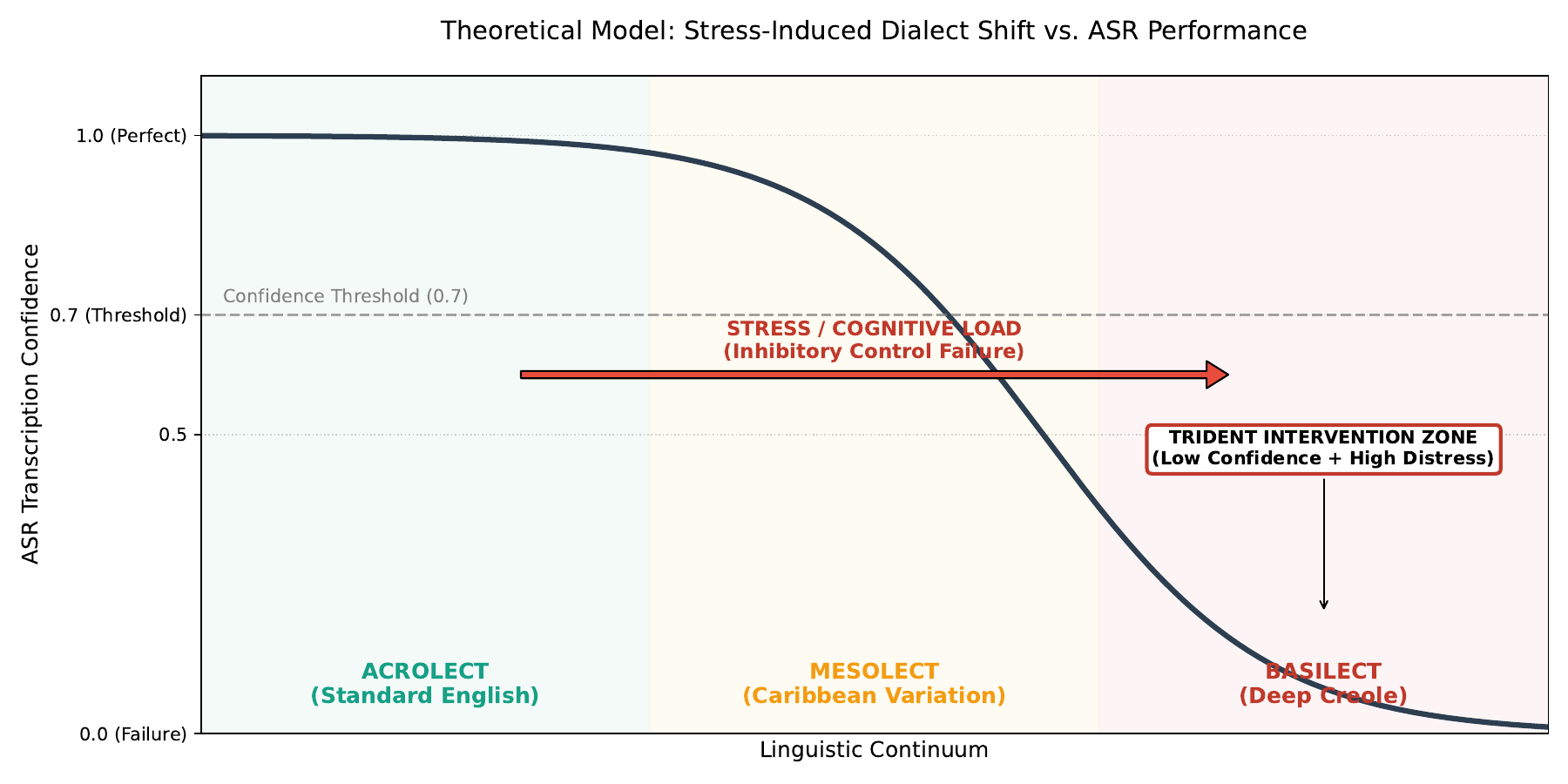}
\caption{The TRIDENT Integration Thesis: Stress-Induced Dialect Shift vs. ASR Performance. The model illustrates the system's theoretical foundation. Under acute stress (red arrow), speakers experience inhibitory control failure, shifting along the continuum from acrolectal (standard) to basilectal (creole) registers. As speech becomes more basilectal, ASR confidence (blue line) degrades below the usable threshold of 0.7. The ``Intervention Zone'' highlights TRIDENT's novel contribution: identifying calls where low transcription confidence coincides with high bio-acoustic distress, thereby converting a technical failure into a high-priority (Q1) routing signal.}
\label{fig:theoretical_model}
\end{figure}

This thesis rests on the psycholinguistic literature establishing that:
\begin{enumerate}
    \item Stress triggers cognitive load effects that impair executive function \cite{gollan2009}
    \item Impaired executive function leads to reduced inhibition of dominant language varieties \cite{green1998}
    \item For Caribbean speakers, dominant varieties include basilectal forms underrepresented in ASR training \cite{patrick1999, madden2025}
    \item Stress simultaneously elevates bio-acoustic markers (F0, intensity) that can be detected independently of speech content \cite{vanpuyvelde2018}
\end{enumerate}

The logical conclusion: when ASR confidence drops and bio-acoustic distress rises, the system has detected a caller in genuine crisis whose speech has shifted beyond standard recognition capabilities. This combination should trigger immediate human review---not because the system has failed, but because it has successfully identified a caller who needs human attention most.

%% file: sections/03_system_architecture.tex
\section{System Architecture}

\begin{figure}[!htb]
\centering
\includegraphics[width=0.9\textwidth]{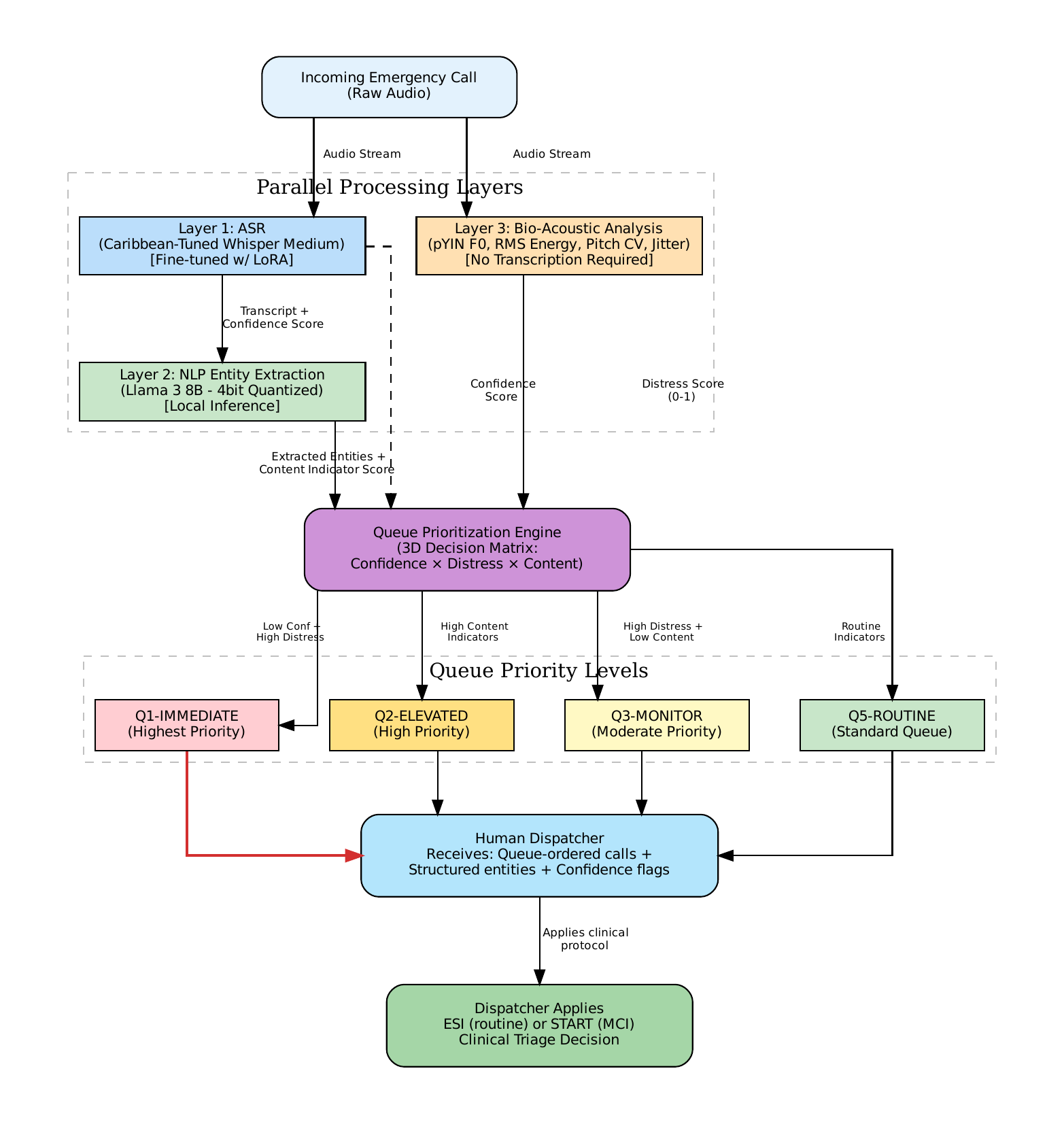}
\caption{The TRIDENT architecture. The system processes raw audio through two parallel streams: (Left) A Caribbean-adapted ASR and NLP pipeline for entity extraction and content analysis, and (Right) a bio-acoustic analysis layer for detecting physiological distress markers. The \textbf{Queue Prioritization Engine} integrates three independent signals---transcription confidence, extracted clinical indicators, and vocal distress---to determine queue position for dispatcher attention. This ensures that (1) calls with low transcription confidence but high vocal distress receive immediate human review, and (2) semantically urgent calls from calm reporters are not delayed due to absent vocal stress markers. The dispatcher then applies established triage protocols (ESI for routine operations, START for mass casualty events) using both TRIDENT's extracted entities and direct audio review.}
\label{fig:system_diagram}
\end{figure}

TRIDENT implements a three-layer dispatcher-support architecture where each component provides independent value while contributing to intelligent queue prioritization. The system does not make clinical triage decisions---those remain with trained dispatchers applying ESI or START protocols---but ensures dispatchers receive the highest-priority calls first along with structured information to support protocol application. Figure~\ref{fig:system_diagram} illustrates the system flow.

\subsection{Design Philosophy: Enabling Protocol Application}

TRIDENT's architecture reflects a core principle: \textbf{AI should empower dispatchers to apply established protocols more effectively, not replace clinical judgment}. Caribbean health ministries have adopted validated triage frameworks, ESI for emergency departments, START for mass casualty incidents, that represent decades of clinical refinement. TRIDENT's role is to solve the \emph{input problem}: ensuring these protocols can be applied equitably to Caribbean-accented callers whose speech current ASR systems fail to transcribe accurately.

Each architectural layer addresses a specific input challenge:

\begin{itemize}
    \item \textbf{Layer 1 (ASR):} Produces transcripts and confidence scores, enabling dispatchers to know when to trust text versus listen directly to audio.
    
    \item \textbf{Layer 2 (NLP):} Extracts structured clinical entities---location, mechanism of injury, breathing status, vulnerable populations---that map directly to ESI/START decision points.
    
    \item \textbf{Layer 3 (Bio-Acoustic):} Detects physiological distress markers that indicate caller crisis state, providing a signal not currently captured by standard protocols but valuable for queue prioritization.
\end{itemize}

The following subsections detail each layer's implementation.

\input{subsections/architecture/asr_layer.tex}
\input{subsections/architecture/nlp_layer.tex}
\input{subsections/architecture/bioacoustic_layer.tex}
\input{subsections/architecture/complementarity.tex}
\input{subsections/architecture/queue_prioritization.tex}

%% file: subsections/architecture/asr_layer.tex
\subsection{Layer 1: Caribbean-Tuned ASR}

The ASR layer employs OpenAI's Whisper Medium fine-tuned with Low-Rank Adaptation (LoRA) on Caribbean broadcast speech. We selected Whisper Medium over Large based on Madden et al.'s \cite{madden2025} scaling law, which demonstrates that domain-specific data yields greater gains than model size for Caribbean varieties. Whisper Medium is also more efficient for Raspberry Pi 5 edge deployment.

\textbf{Fine-tuning Configuration:}
\begin{itemize}
    \item Base model: openai/whisper-medium
    \item Adaptation: LoRA (rank=16, alpha=32)
    \item Training data: BBC Caribbean broadcast corpus ($\sim$28,000 clips)
    \item Trainable parameters: $\sim$0.5\% of total model
\end{itemize}

\textbf{Confidence Scoring:} The system computes utterance-level confidence as the mean log-probability across all decoded tokens, normalized to 0-1:

\begin{equation}
\text{confidence} = \exp\left(\frac{1}{N}\sum_{i=1}^{N} \log P(t_i | t_1 \ldots t_{i-1}, \text{audio})\right)
\end{equation}

We use utterance-level rather than token-level confidence because emergency triage requires holistic assessment of transcription reliability. The low confidence threshold is set at 0.7 based on initial calibration.

%% file: subsections/architecture/nlp_layer.tex
\subsection{Layer 2: Local NLP Entity Extraction}

When ASR produces usable transcription (confidence $\geq$ 0.7), the NLP layer extracts structured emergency information using Llama 3 8B running locally via Ollama. The extraction schema targets entity types that map directly to ESI and START triage protocol decision points.

\subsubsection{Entity Extraction Schema}

The schema targets four entity categories:
\begin{itemize}
    \item \textbf{LOCATION:} Street addresses, landmarks, geographic references
    \item \textbf{MECHANISM/HAZARD:} Emergency type (fire, flood, medical, violence, traffic)
    \item \textbf{CLINICAL INDICATORS:} Breathing status, consciousness, bleeding, mobility
    \item \textbf{SCALE:} Number of people involved, vulnerable populations
\end{itemize}

\subsubsection{Mapping to Triage Protocols}

TRIDENT entities support ESI and START protocol application. For ESI, extracted entities inform the four decision points: Point A (lifesaving intervention) captures "not breathing," "choking," "unresponsive"; Point B (high-risk situation) captures mechanism of injury and altered status; Point C (resource needs) uses hazard type and complexity; Point D (vital signs) uses reported vitals and distress indicators \cite{esi_handbook}.

For mass casualty events using START, entities support rapid sorting: GREEN captures "walking," "minor injuries"; YELLOW captures "injured but stable," "conscious"; RED captures "trapped," "not breathing," "heavy bleeding"; BLACK captures cessation indicators.

\begin{table}[ht]
\centering
\small
\begin{tabular}{@{}p{2.5cm}p{4cm}p{5.5cm}@{}}
\toprule
\textbf{Protocol} & \textbf{Decision Point} & \textbf{Example Extraction Target} \\ \midrule
ESI Level 1 & Immediate lifesaving intervention? & ``not breathing,'' ``choking,'' ``heavy bleeding,'' ``unresponsive'' \\[0.5em]
START RED & Not walking, breathing issues & ``trapped,'' ``not breathing,'' ``unresponsive,'' ``heavy bleeding'' \\ \bottomrule
\end{tabular}
\caption{Example entity extraction targets supporting ESI and START protocols. Full protocol mappings detailed in extended version.}
\label{tab:protocol_mapping_simplified}
\end{table}

\subsubsection{Handling Garbled Input}

The NLP layer handles low-quality transcriptions through confidence-aware prompting. When ASR confidence is below 0.7, the system instructs the LLM to mark uncertain extractions, avoid hallucination, prioritize location extraction, and note phonetically similar alternatives. When confidence is very low ($<$0.4), minimal structured output is produced and the call is flagged for immediate human review.

\subsubsection{Content Indicator Scoring}

The NLP layer computes a \textbf{Content Indicator Score} ($S_c \in [0,100]$) quantifying urgency implied by semantic content, independent of how the caller sounds. This addresses a critical gap: a trained first responder may report a mass casualty event calmly, producing low bio-acoustic distress despite extremely urgent content. Without content analysis, such calls would be deprioritized.

Rather than keyword matching, we leverage the LLM's semantic understanding to classify transcript content. This approach handles Caribbean creole variants (``mi granmodda drop dung an she nah move'' conveys the same urgency as ``my grandmother collapsed and she's not moving''), negation, and indirect references.

The LLM outputs structured classifications:
\begin{verbatim}
{
  "hazard_category": "violent_crime" | "medical" | "fire" |
                     "flood" | "traffic" | "infrastructure" | "other",
  "life_threat_level": "imminent" | "potential" | "none",
  "vulnerable_population": true | false,
  "situation_status": "escalating" | "stable" | "resolved",
  "persons_affected": <integer>
}
\end{verbatim}

A deterministic function maps classifications to the score:
\begin{equation}
S_c = \min\left(100,\ S_{\text{hazard}} + S_{\text{threat}} + S_{\text{vuln}} + S_{\text{scale}}\right)
\label{eq:content_severity}
\end{equation}

\textbf{Scoring components:} Hazard category weights range from 30 (violent crime) to 5 (other). Life-threat level contributes +30 (imminent), +15 (potential), or +0 (none). Vulnerable population adds +15. Scale combines persons affected (+5 per person, capped at +20) and escalation status (+10 if escalating).

\textbf{Example calculations:}

\begin{table}[ht]
\centering
\small
\begin{tabular}{@{}p{0.40\textwidth}p{0.35\textwidth}c@{}}
\toprule
\textbf{Transcript} & \textbf{Classification} & \textbf{$S_c$} \\ \midrule
``Pothole on Nelson Street'' &
infrastructure, none, false, stable, 0 & 10 \\[0.5em]
``House fire, spreading to neighbor's yard'' &
fire, potential, false, escalating, 0 & 50 \\[0.5em]
``Pickney dem trap inna di fire'' &
fire, imminent, true, stable, 2+ & 80 \\ \bottomrule
\end{tabular}
\caption{Content indicator scoring via LLM classification. Semantic understanding captures urgency from Caribbean creole variants. High scores elevate queue priority; clinical triage remains with dispatchers.}
\label{tab:severity_examples}
\end{table}

The Content Indicator Score feeds into queue prioritization (Section~\ref{sec:queue_prioritization}), ensuring semantically urgent calls reach dispatchers promptly even when vocal distress markers are absent. Weights are tunable parameters that should be calibrated with local emergency services to reflect institutional priorities and regional hazard profiles.

%% file: subsections/architecture/bioacoustic_layer.tex
\subsection{Layer 3: Bio-Acoustic Distress Detection}

The bio-acoustic layer operates on raw audio, independent of ASR success, extracting features correlated with psychological distress. Based on the vocal stress literature \cite{schmalz2025, vanpuyvelde2018, veiga2025}, we focus on features that capture physiological arousal through vocal production changes.

\subsubsection{Feature Extraction}

Using librosa, we extract the following acoustic features:

\begin{enumerate}
    \item \textbf{Fundamental Frequency (F0):} Mean pitch extracted via autocorrelation method
    \begin{itemize}
        \item Typical baseline: 85--180 Hz (male), 165--255 Hz (female) \cite{titze1989}
        \item Stress indicator: Elevation above speaker baseline
    \end{itemize}

    \item \textbf{F0 Coefficient of Variation (CV):} Pitch instability measure
    \begin{itemize}
        \item Computed as $CV = \sigma_{F0} / \mu_{F0}$
        \item Normalizes for baseline differences across speakers
        \item Stress indicator: $CV > 0.3$ suggests vocal instability
    \end{itemize}

    \item \textbf{Energy (RMS amplitude):} Mean intensity across utterance
    \begin{itemize}
        \item Normalized to 0--1 scale relative to recording gain
        \item Stress indicator: Elevated intensity during distress vocalizations
    \end{itemize}

    \item \textbf{Jitter:} Cycle-to-cycle variation in F0 period
    \begin{itemize}
        \item Relatively independent of prosodic patterns \cite{vanpuyvelde2018}
        \item Pathology threshold: $>$1.04\% \cite{boersma2013}
    \end{itemize}
\end{enumerate}

\subsubsection{Distress Score Calculation}

The distress score combines multiple acoustic indicators into a composite metric. We weight features according to their documented reliability and sex-independence:

\begin{align}
D &= w_{\text{pitch}} \cdot P + w_{\text{var}} \cdot V + w_{\text{energy}} \cdot E + w_{\text{jitter}} \cdot J
\label{eq:distress}
\end{align}

\noindent where:

\noindent The pitch elevation component now uses sex-adaptive parameters:
\begin{equation}
P = \min\left(1.0, \max\left(0, \frac{\bar{F_0} - B}{R}\right)\right) \quad \text{(pitch elevation)}
\end{equation}

\noindent where $(B, R)$ adapts based on estimated speaker sex:
\begin{equation}
(B, R) = \begin{cases}
    (120, 80) & \text{if } \bar{F_0}^{(\text{init})} < 165 \text{ Hz (estimated male)}\\
    (200, 100) & \text{otherwise (estimated female)}
\end{cases}
\end{equation}

The baseline $B$ and range $R$ parameters adapt based on a heuristic sex estimation from the initial 3 seconds of speech. A male speaker at 170 Hz (stressed) now contributes $P = (170-120)/80 = 0.625$ rather than the previous formulation's 0.0, addressing the male pitch penalty.

\noindent The remaining components are:
\begin{align}
V &= \min\left(1.0, \frac{CV_{F0}}{0.5}\right) & \text{(pitch instability)} \\
E &= \min\left(1.0, \frac{\bar{E}}{0.1}\right) & \text{(energy)} \\
J &= \min\left(1.0, \frac{\text{jitter}}{0.02}\right) & \text{(perturbation)}
\end{align}

The weights reflect relative reliability from the literature:
\begin{itemize}
    \item $w_{\text{pitch}} = 0.30$ --- F0 elevation is the most consistent stress marker but is sex-dependent
    \item $w_{\text{var}} = 0.35$ --- F0 coefficient of variation is sex-normalized and robust
    \item $w_{\text{energy}} = 0.20$ --- intensity elevation accompanies distress
    \item $w_{\text{jitter}} = 0.15$ --- perturbation measures are prosody-independent
\end{itemize}

\subsubsection{Threshold Classification}

\begin{itemize}
    \item \textbf{High Distress:} $D > 0.5$
    \item \textbf{Low Distress:} $D \leq 0.5$
\end{itemize}

These thresholds are calibrated against Van Puyvelde et al.'s \cite{vanpuyvelde2018} findings on vocal markers in emergency versus baseline speech.

\textbf{Note on sex differences:} The distress score prioritizes sex-normalized features (CV, jitter) over absolute F0 elevation to mitigate the substantial baseline differences between male (85--175 Hz) and female (165--270 Hz) speakers. See Section~\ref{sec:sex_limitations} for detailed discussion of remaining bias risks.

%% file: subsections/architecture/complementarity.tex
\subsection{The Complementarity Principle}

The theoretical foundation for our multi-layer design rests on what we term the \textbf{Complementarity Principle}: the three signal dimensions capture distinct failure modes and urgency indicators that compensate for each other's blind spots, ensuring dispatchers receive the most critical calls first regardless of which individual signal might fail.

\textbf{Dimension 1: Transcription Confidence.} The conditions that degrade ASR performance (high stress, code-switching to basilect, environmental noise) are precisely the conditions that often accompany genuine emergencies. Low confidence is not merely a technical limitation to be hidden---it correlates with caller distress and should elevate queue priority while flagging the call for direct audio review.

\textbf{Dimension 2: Content Indicators.} Semantic analysis of transcript content captures urgency that vocal characteristics may miss. Trained professionals, repeat callers, and composed bystanders often report critical emergencies without elevated vocal stress---their calm delivery masks the urgency that only content analysis reveals. When transcription confidence is high, extracted entities map directly to ESI/START decision points.

\textbf{Dimension 3: Bio-Acoustic Distress.} Vocal stress markers (elevated pitch, intensity, instability) provide a parallel assessment channel that operates on raw audio, independent of transcription success. A caller whose speech is entirely unintelligible to ASR will still produce detectable distress signals. This dimension captures information not currently used by ESI or START protocols, representing TRIDENT's novel contribution to dispatcher awareness.

This creates a robust prioritization space with complementary coverage:

\textbf{Dimensional ordering.} The three dimensions are evaluated in deliberate sequence: \emph{Confidence}, \emph{Content}, \emph{Concern}. This ordering reflects operational logic: (1) \emph{Can we understand the caller?}---ASR confidence determines whether transcription is reliable enough for downstream analysis; (2) \emph{What is being reported?}---semantic content establishes the substance of the emergency; (3) \emph{How distressed does the caller sound?}---bio-acoustic indicators validate and can elevate priority, but do not override content. This sequence ensures that a composed professional reporting a mass casualty event receives appropriate priority based on content, while a highly distressed caller reporting a minor issue is not over-prioritized based on vocal expression alone.

\begin{itemize}
    \item \textbf{High Confidence + Low Content + Low Concern:} Routine call; dispatcher applies ESI using extracted entities at normal pace
    
    \item \textbf{High Confidence + High Content + Low Concern:} The composed reporter---urgent content from a calm caller requires elevated queue position; dispatcher reviews entities and applies ESI, likely assigning ESI-2 or ESI-3
    
    \item \textbf{High Confidence + Low Content + High Concern:} Anxious caller, possibly minor issue---dispatcher assesses whether distress reflects emergency or anxiety
    
    \item \textbf{High Confidence + High Content + High Concern:} All signals aligned; immediate queue position for rapid ESI/START application
    
    \item \textbf{Low Confidence + Low Content + Low Concern:} Likely technical issue; dispatcher reviews audio quality before processing
    
    \item \textbf{Low Confidence + High Content + Low Concern:} Garbled but fragments suggest urgency---elevated priority; dispatcher listens directly
    
    \item \textbf{Low Confidence + Low Content + High Concern:} Distressed caller with unintelligible speech---immediate priority; dispatcher listens and applies protocol based on direct assessment
    
    \item \textbf{Low Confidence + High Content + High Concern:} Maximum queue priority---all indicators suggest crisis; immediate dispatcher attention
\end{itemize}

Two cells represent our key insights. The \textbf{High Confidence + High Content + Low Concern} cell captures callers whose semantic content demands urgent attention despite calm delivery: the trained first responder, medical professional, or composed bystander whose measured voice belies the severity of their report. The \textbf{Low Confidence + Low Content + High Concern} cases capture the complementary pattern---callers in crisis whose speech has shifted toward basilectal registers, where ASR failure combined with vocal stress becomes valuable prioritization information rather than system failure.

Together, these insights ensure that neither semantic nor paralinguistic signals alone determine queue position---and that clinical triage decisions remain with trained dispatchers who can assess the full context of each call.

%% file: subsections/architecture/queue_prioritization.tex
\subsection{Queue Prioritization Engine}
\label{sec:queue_prioritization}

The Queue Prioritization Engine integrates three independent signals to determine the order in which calls receive dispatcher attention. \textbf{Critically, this system determines queue position, not clinical triage category.} Clinical triage---assigning ESI levels 1--5 or START colors (RED/YELLOW/GREEN/BLACK)---remains the responsibility of trained dispatchers applying Ministry of Health protocols.

The prioritization logic ensures that:
\begin{enumerate}
    \item Callers most likely to need immediate intervention reach dispatchers first
    \item Dispatchers receive structured information to support rapid protocol application
    \item Calls with unreliable transcriptions are flagged for direct audio review
\end{enumerate}

\subsubsection{Three-Dimensional Prioritization Space}

Each call is mapped to a point in prioritization space defined by:

\begin{itemize}
    \item \textbf{Transcription Confidence} ($C$): High ($\geq 0.7$) or Low ($< 0.7$)
    \item \textbf{Content Indicators} ($S_c$): High ($\geq 50$) or Low ($< 50$)
    \item \textbf{Bio-Acoustic Distress} ($D$): High ($> 0.5$) or Low ($\leq 0.5$)
\end{itemize}

The $2 \times 2 \times 2$ combination yields eight queue priority cells, shown in Table~\ref{tab:queue_matrix_3d}.

\begin{table}[ht]
\centering
\small
\begin{tabular}{@{}ccclp{4.8cm}@{}}
\toprule
\textbf{Confidence} & \textbf{Content} & \textbf{Concern} & \textbf{Queue} & \textbf{Dispatcher Action} \\ \midrule
High & Low & Low & \textbf{Q5-ROUTINE} & Apply ESI using extracted entities \\
High & High & Low & \textbf{Q2-ELEVATED} & Priority review; calm reporter, urgent content$^*$ \\
High & Low & High & \textbf{Q3-MONITOR} & Review for anxiety vs. emergency \\
High & High & High & \textbf{Q1-IMMEDIATE} & Immediate attention; apply ESI/START \\
Low & Low & Low & \textbf{Q5-REVIEW} & Check audio quality; possible technical issue \\
Low & High & Low & \textbf{Q2-ELEVATED} & Listen to audio; fragments suggest urgency \\
Low & Low & High & \textbf{Q1-IMMEDIATE} & Priority audio review; possible dialect shift$^\dagger$ \\
Low & High & High & \textbf{Q1-IMMEDIATE} & Highest priority; all indicators elevated \\ \bottomrule
\end{tabular}
\caption{Three-dimensional queue prioritization matrix. $^*$Addresses trained responder/composed bystander scenario. $^\dagger$Preserves core insight: low ASR confidence + high vocal concern may indicate stress-induced basilectal shift requiring human ears.}
\label{tab:queue_matrix_3d}
\end{table}

\subsubsection{Queue Priority Levels}

\begin{description}
    \item[Q1-IMMEDIATE:] Top of queue. Dispatcher reviews within seconds. System flags call for potential crisis requiring direct audio assessment.
    
    \item[Q2-ELEVATED:] High priority queue. Dispatcher attention within 1--2 minutes. Extracted entities displayed prominently to support rapid ESI/START application.
    
    \item[Q3-MONITOR:] Moderate priority. May indicate anxious caller with non-urgent situation. Dispatcher assesses and de-escalates if appropriate.
    
    \item[Q5-ROUTINE:] Standard queue. Extracted entities available; dispatcher applies ESI at normal pace.
    
    \item[Q5-REVIEW:] Standard queue but flagged for audio quality check. May indicate technical issues rather than emergency content.
\end{description}

\textbf{Note on Q4:} The current matrix does not produce a Q4 outcome. Future refinement with real operational data may identify scenarios warranting an intermediate priority level. A theoretical case: High Confidence + Low Content + Moderate Concern (anxious caller, minor issue).

\subsubsection{Relationship to Clinical Triage Protocols}

Table~\ref{tab:protocol_mapping} illustrates how TRIDENT's queue prioritization relates to---but does not replace---clinical triage protocols.

\begin{table}[ht]
\centering
\small
\begin{tabular}{@{}p{2.5cm}p{5cm}p{5cm}@{}}
\toprule
\textbf{TRIDENT Output} & \textbf{Dispatcher Action} & \textbf{Protocol Application} \\ \midrule
Q1-IMMEDIATE & Immediate audio review; assess caller state & Dispatcher determines ESI-1/2 or START-RED based on clinical assessment \\[0.5em]
Q2-ELEVATED & Review extracted entities; listen if uncertain & Dispatcher applies ESI using structured data; may be ESI-2 through ESI-4 \\[0.5em]
Q3-MONITOR & Assess distress source; de-escalate if needed & Often ESI-4/5 after dispatcher determines no emergency \\[0.5em]
Q5-ROUTINE/REVIEW & Process normally using extracted metadata & Full ESI protocol application; typically ESI-3 through ESI-5 \\ \bottomrule
\end{tabular}
\caption{TRIDENT queue priority does not determine clinical triage level. Dispatchers apply ESI or START protocols after reviewing TRIDENT's structured outputs and/or call audio.}
\label{tab:protocol_mapping}
\end{table}

\subsubsection{Dispatcher Interface}

\begin{figure}[!htb]
\centering
\includegraphics[width=0.85\textwidth]{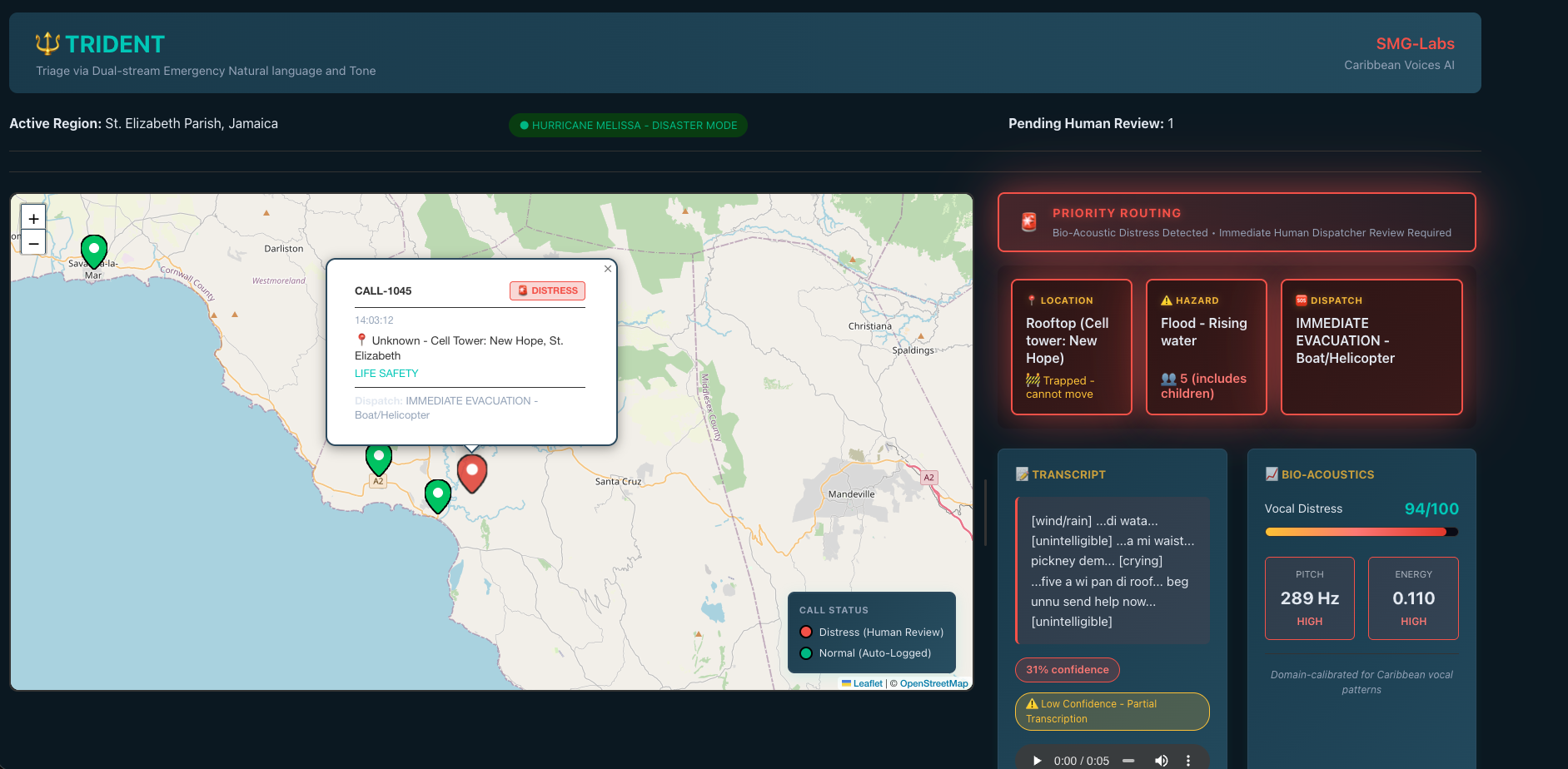}
\caption{Dispatcher interface for a high-priority scenario (Q1-IMMEDIATE). Elevated distress markers combined with low transcription confidence trigger immediate queue placement. The interface prominently recommends audio review and displays partial entity extraction with uncertainty markers. The dispatcher will listen directly and apply ESI or START protocol based on their clinical assessment.}
\label{fig:ui_high_risk}
\end{figure}

Figure~\ref{fig:ui_high_risk} illustrates the dispatcher interface for a high-priority scenario. The interface presents:

\begin{itemize}
    \item Queue priority level with visual urgency coding
    \item Transcription confidence (with recommendation to review audio if low)
    \item Extracted clinical entities mapped to ESI/START decision points
    \item Bio-acoustic distress indicators
    \item One-click access to call audio for direct assessment
\end{itemize}

%% file: sections/05_deployment.tex
\section{Deployment Considerations}

\subsection{Operational Context: Supporting Protocol Application}

TRIDENT integrates with existing emergency dispatch workflows to support standardized triage protocols---ESI for routine operations, START for mass casualty incidents. \textbf{Day-to-day (ESI context):} TRIDENT processes incoming calls to extract structured entities (location, mechanism, clinical indicators) and assigns queue priority. Dispatchers apply ESI to determine clinical acuity level (1--5) and appropriate response. \textbf{Mass casualty events (START context):} During hurricanes or earthquakes, TRIDENT's queue prioritization manages call surges when volume exceeds dispatcher capacity, enabling rapid caller sorting even when transcription quality degrades. \textbf{Key principle:} TRIDENT determines which calls dispatchers see first and what structured information they receive; clinical triage decisions remain with trained professionals applying Ministry of Health protocols.

\subsection{Primary Deployment: Surge Queue Prioritization}

TRIDENT's greatest value emerges during \textbf{disaster surge conditions}---hurricanes, earthquakes, floods---when call volume exceeds dispatcher capacity and callers must wait in queue. TRIDENT's processing latency (45--60 seconds on edge hardware) precludes real-time transcription, but surge queues provide ideal operational context.

\textbf{Operational flow:}
\begin{enumerate}
    \item Caller dials emergency services; all dispatchers engaged
    \item Caller enters queue and hears automated message requesting description
    \item Caller provides initial statement (15--30 seconds)
    \item TRIDENT processes audio while caller waits (45--60 seconds)
    \item Queue reordered by priority (Q1-IMMEDIATE through Q5-ROUTINE)
    \item Highest-priority call routes first when dispatcher becomes available
    \item Dispatcher receives transcription, extracted entities, and distress indicators to support ESI/START application
\end{enumerate}

\textbf{Why this context maximizes value:} Calls are waiting regardless---TRIDENT uses wait time productively. Queue prioritization ensures most critical callers reach dispatchers first. Extracted entities enable faster protocol application. Low ASR confidence flags alert dispatchers to potential dialect shift or audio quality issues before engagement.

This deployment model represents TRIDENT's primary design target. Caribbean emergency services face predictable annual surge events (hurricane season, June--November) where this capability would directly impact response effectiveness.

\subsection{Early Exit for Critical Cases}

To provide faster routing for clearly distressed callers, the system implements early exit when:
\begin{enumerate}
    \item \textbf{High Distress + Low Confidence:} If $D > 0.8$ and $C < 0.4$, route immediately to Q1-IMMEDIATE. This captures callers exhibiting extreme vocal stress whose speech has likely shifted to basilectal registers.

    \item \textbf{Extreme Distress:} If $D > 0.9$ regardless of confidence, route to Q1-IMMEDIATE.
\end{enumerate}

Under early exit, ASR and bio-acoustics complete in approximately 12 seconds (with bio-acoustic extraction parallel to transcription), reducing Time-to-Q1 from 55 seconds to 12 seconds for clearly distressed callers---a critical improvement for surge queue scenarios.

\subsection{Offline Operation}

All components operate without internet connectivity: Whisper model weights and Llama 3 stored locally, bio-acoustic analysis uses standard signal processing libraries, and queue logic implemented in local Python. This enables deployment at emergency coordination centers that may lose connectivity during disasters while maintaining local power (generator/battery backup). Offline capability ensures TRIDENT can support ESI/START protocol application precisely when infrastructure degradation makes accurate call processing most difficult.

\subsection{Integration with Existing Dispatch Systems}

TRIDENT operates as a \textbf{pre-processing layer} integrating with existing Computer-Aided Dispatch (CAD) systems. The system accepts audio streams, processes them through the three-layer architecture, and outputs structured data packages (queue priority, transcription with confidence, extracted entities, distress indicators) to CAD systems. Dispatchers receive calls in priority order and apply ESI or START protocols using TRIDENT's structured data and/or direct audio review. This requires no changes to clinical protocols---only familiarization with TRIDENT's output format.

\subsection{Hardware Requirements}

The complete system deploys on Raspberry Pi 5 (8GB RAM) or equivalent edge hardware:

\begin{table}[ht]
\centering
\begin{tabular}{@{}llll@{}}
\toprule
\textbf{Component} & \textbf{Model} & \textbf{Size} & \textbf{Inference Speed} \\ \midrule
ASR & Whisper Medium (INT4) & $\sim$400MB & $\sim$10s per 30s audio \\
NLP & Llama 3 8B (4-bit) & $\sim$4GB & 2-5 tokens/sec \\
Bio-acoustic & librosa + numpy & $<$50MB & Real-time \\ \bottomrule
\end{tabular}
\caption{Hardware requirements for edge deployment}
\label{tab:hardware}
\end{table}

Total system footprint: $\sim$4.5GB, well within Raspberry Pi 5 8GB capacity.

%% file: sections/06_limitations.tex
\section{Limitations and Future Work}

\subsection{Current Limitations}

\textbf{Validation gap (most critical).} This paper presents an architectural framework with theoretical grounding but limited empirical validation on real emergency calls. Performance claims are based on component evaluations and related literature rather than end-to-end system testing. The three-dimensional queue prioritization matrix has not been validated against expert dispatcher judgments.

\textbf{Protocol integration.} While TRIDENT is framed as supporting ESI and START protocols, the entity extraction schema and queue prioritization logic were developed independently of clinical stakeholder input. Full Ministry of Health integration requires validation that extracted entities map correctly to ESI decision points and that queue priorities align with operational workflows.

\textbf{Training data constraints.} Caribbean emergency speech corpora do not exist. ASR fine-tuning was performed on broadcast speech, which differs from emergency call acoustics in noise profiles, emotional content, and register distribution.

\textbf{Sex differences in F0 baseline.}
\label{sec:sex_limitations}
Fundamental frequency is sexually dimorphic: male voices typically range 85--175 Hz while female voices range 165--270 Hz \cite{titze1989, traunmuller1995}. We mitigate this by prioritizing sex-normalized features (F0 coefficient of variation, jitter) over absolute F0 elevation in distress score calculation. Research confirms that stress manifests with ``striking parallels in men and women'' \cite{pisanski2018}---both sexes show increased pitch mean and variation under acute stress. However, residual bias risks remain: relaxed female speakers near upper baseline may contribute to elevated distress scores, while stressed male speakers with naturally low F0 may not contribute sufficiently. A validation study with sex-stratified analysis on Caribbean emergency calls is essential to calibrate population-appropriate thresholds and confirm normalized measures maintain sensitivity across demographics.

\textbf{Content indicator classification.} The Content Indicator Score depends on LLM classification quality. Caribbean creole expressions not well-represented in training data may be misclassified. Empirical evaluation of classification accuracy on Caribbean transcripts is needed, particularly for false negatives that could delay critical calls.

\textbf{Single-speaker assumption.} Multi-party calls are not handled. Speaker changes mid-call could confuse bio-acoustic analysis and entity extraction.

\textbf{Threshold sensitivity.} Multiple thresholds (ASR confidence 0.7, distress 0.5, content indicators 50) were selected based on literature but have not been rigorously optimized. Sensitivity analysis examining precision-recall tradeoffs is needed.

\subsection{Future Work}

\textbf{Clinical stakeholder collaboration.} Partnership with Caribbean emergency services to validate TRIDENT's utility in real dispatch workflows, including observation studies of current ESI/START challenges, dispatcher feedback on extracted entity usefulness, and iterative schema refinement based on clinical input.

\textbf{Caribbean Emergency Speech Corpus.} A dedicated corpus combining Caribbean-accented speech with emergency domain content and stress annotations is critical. We are exploring \textit{VoicefallJA}, a gamified speech elicitation platform designed to collect stressed Caribbean speech through game-induced cognitive load rather than acted performance. The Progressive Web App targets 100--300 speakers via church network distribution, with Q2--Q3 2026 data collection. However, game-induced stress differs fundamentally from genuine emergency distress; this approach should be viewed as a stepping stone toward real-call annotation under appropriate ethical frameworks, not a replacement.

\textbf{Empirical validation.} End-to-end evaluation with emergency dispatch professionals assessing whether TRIDENT's queue prioritization aligns with expert judgment, including sex-stratified analysis of bio-acoustic accuracy and entity extraction accuracy on Caribbean creole transcripts.

\textbf{Ablation studies.} Quantifying the marginal contribution of each architectural component (bio-acoustic analysis, content indicators, Caribbean-tuned ASR).

\textbf{Sex-adaptive distress detection.} Implementing within-call F0 change detection rather than absolute thresholds, and ensemble approaches combining multiple normalization strategies.

%% file: sections/07_conclusion.tex
\section{Conclusion}

TRIDENT presents a dispatcher-support architecture that ensures Caribbean-accented emergency callers receive equitable access to ESI and START triage protocols. By combining accent-adapted speech recognition, local NLP entity extraction, and bio-acoustic distress detection, the system empowers dispatchers to apply established protocols even when automated transcription fails.

The architecture operationalizes two complementary insights established in Section~\ref{sec:key_insights}: that ASR uncertainty combined with vocal distress signals priority callers requiring human attention, and that calm delivery of urgent content must not delay dispatcher response. These insights drive the three-dimensional queue prioritization matrix that routes calls based on confidence, content, and concern signals.

Critically, TRIDENT respects the clinical authority of established protocols. The system determines which calls dispatchers see first and provides structured information to support rapid protocol application---but triage decisions remain with trained human professionals. This design philosophy reflects a broader principle for emergency AI: technology should empower human expertise, not attempt to replace it.

We hope this architectural framework contributes to more equitable emergency services---not just for Caribbean populations, but for the billions of speakers worldwide whose accents and dialects remain underserved by current speech technology. When a caller dials for help, the system that answers should understand them. TRIDENT is a step toward that goal.

%% file: appendices/implementation.tex
\section{Implementation Details}

\textbf{Repository:} \url{https://github.com/smg-labs/project-filter} \emph{(to be made public upon acceptance)}

\textbf{Dependencies:}
\begin{itemize}
    \item Python 3.11+
    \item openai-whisper
    \item transformers, peft (LoRA fine-tuning)
    \item ollama (Llama 3 serving)
    \item librosa (audio feature extraction)
    \item jiwer (WER evaluation)
\end{itemize}

\textbf{Hardware requirements:}
\begin{itemize}
    \item Training: NVIDIA GPU with 16GB+ VRAM recommended
    \item Inference: CPU-only operation supported; 8GB RAM minimum
\end{itemize}

%% file: appendices/acknowledgments.tex
\section{Acknowledgments}

This work emerged from the Caribbean Voices AI Hackathon, organized by the UWI AI Innovation Centre and hosted on Zindi. We thank sponsors CIBC, Infolytics, and DataAxis for their support. The competition's BBC Caribbean speech corpus motivated this architectural framework. We also thank Dr. Sikopo Nyambe-Galbraith for feedback on the research.

We also acknowledge the use of Google's Gemini 3.0 Pro when brainstorming the ideas for the paper, conducting deep research, and the generation of the figures in this paper. Anthropic's Claude Opus 4.5 was used for the editing and proofreading of the paper.